\documentclass[10pt,twocolumn,letterpaper]{article}

\usepackage[pagenumbers]{cvpr} 

\usepackage{graphicx}
\usepackage{amsmath}
\usepackage{amssymb}
\usepackage{booktabs}

\usepackage{multirow} 
\usepackage{makecell} 
\usepackage{soul}
\usepackage{amsthm}

\newtheorem{proposition}{Proposition}
\newtheorem{theorem}{Theorem}
\newtheorem*{remark}{Remark}

\usepackage[accsupp]{axessibility}  

\usepackage[pagebackref,breaklinks,colorlinks]{hyperref}

\usepackage[capitalize]{cleveref}
\crefname{section}{Sec.}{Secs.}
\Crefname{section}{Section}{Sections}
\Crefname{table}{Table}{Tables}
\crefname{table}{Tab.}{Tabs.}

\def\cvprPaperID{8843} 
\def\confName{CVPR}
\def\confYear{2023}

\begin{document}

\title{Shape-Erased Feature Learning for Visible-Infrared Person Re-Identification}

\author{Jiawei Feng$^{1}$  \qquad\quad Ancong Wu$^{1}$\thanks{Corresponding author} \qquad \quad Wei-Shi Zheng$^{1,2,3}$\\
$^1$School of Computer Science and Engineering, Sun Yat-sen University, China\\
$^2$Key Laboratory of Machine Intelligence and Advanced Computing, Ministry of Education, China \\
$^3$Guangdong Key Laboratory of Information Security Technology, China \\
{\tt\small fengjw3@mail2.sysu.edu.cn, wuanc@mail.sysu.edu.cn, wszheng@ieee.org}
}
\maketitle

\begin{abstract}
Due to the modality gap between visible and infrared images with high visual ambiguity, learning \textbf{diverse} modality-shared semantic concepts for visible-infrared person re-identification (VI-ReID) remains a challenging problem.
Body shape is one of the significant modality-shared cues for VI-ReID.
To dig more diverse {modality-shared} cues, we expect that erasing body-shape-related semantic concepts in the learned features can force the ReID model to extract more and other modality-shared features for identification. 
To this end, we propose shape-erased feature learning paradigm that decorrelates modality-shared features in two orthogonal subspaces. 
Jointly learning shape-related feature in one subspace and shape-erased features in the orthogonal complement achieves a conditional mutual information maximization between shape-erased feature and identity discarding body shape information, thus enhancing the diversity of the learned representation explicitly.
Extensive experiments on SYSU-MM01, RegDB, and HITSZ-VCM datasets demonstrate the effectiveness of our method.\footnote{Code will be available at \url{https://github.com/jiawei151/SGIEL_VIReID}.}
\end{abstract}

\section{Introduction}
\label{sec:intro}
Recently, person re-identification (ReID) for pedestrian matching in non-overlapping camera views has experienced fast development. However, ReID is still challenging when people appear both in the daytime and in low-light situations where only infrared cameras can clearly capture their appearances, raising the task of visible-infrared ReID (VI-ReID).
Many remarkable works \cite{sysumm01,gan,duallevel,xmodal,twostream,sdl} have been witnessed in the field of VI-ReID. 
For realistic scenarios, discovering\textit{ rich and diverse} modality-shared semantic concepts usually helps to improve the effectiveness of VI-ReID \cite{nuances,paenet}. So far, \textit{diverse} modality-shared feature learning remains challenging.
\begin{figure}[t]
  \centering
  \includegraphics[width=\linewidth]{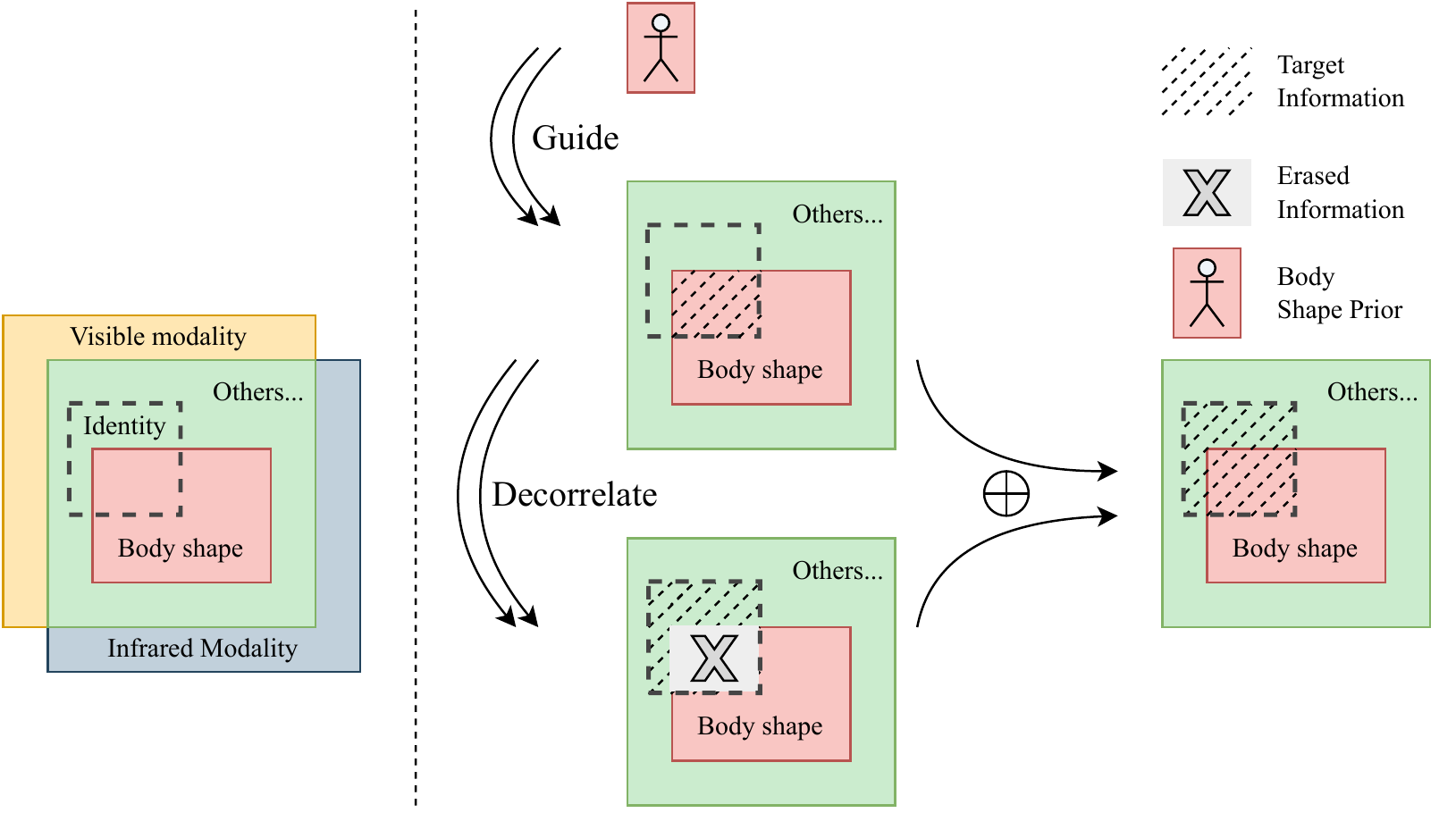}
  \caption{An illustration of our motivation on VI-ReID. It is assumed that body shape information and identity-related modality-shared information (presented in dashed box) are partially overlapped with each other. To make extracted features more \textit{diverse}, we propose shape-erased feature learning paradigm that decomposes the representation into shape-related feature and shape-erased one.
  Learning shape-erased feature drives the model to discover richer modality-shared semantic concepts other than body shape.
  }
  \vspace{-0.2cm}
  \label{fig:1}
\end{figure}


Among the cues for VI-ReID, we can identify pedestrians by their body shapes in many situations, for it contains modality-invariant information and also robust to light changes. Nevertheless,
body shape is not the only or a sufficient semantic concept that interprets the identity of a person. It may be hard in some situations to tell the difference only depending on the body shape, but we can still distinguish them by other semantic concepts, such as their belongings, hairstyles or face structures.
Inspired by this, we illustrate an information theoretic measure between visible and infrared modality as a Venn diagram on the left of the dashed line in Fig.~\ref{fig:1}. 
It is assumed that body shape (presented in red) and identity-related modality-shared information (presented in dashed box) are partially overlapped with each other. Note that \textit{partially} is also due to there exists identity-unrelated information contained in body shape map, \eg, human pose.
This partially overlapped assumption indicates that the target information for VI-ReID, which is identity-related and modality-shared, can be divided into two independent components that are related and unrelated to body shape. 

Based on the above observation and assumption, to dig more diverse modality-shared cues for VI-ReID, we expect to erase the body-shape-related semantic concepts in the features to force the VI-ReID model to extract more and other modality-shared features for identification. 
As illustrated on the right of the dashed line in Fig.~\ref{fig:1}, the \textit{shape-erased} feature is decorrelated from the \textit{shape-related} feature to simultaneously discover shape-unrelated knowledge, while \textit{shape-related} feature can be explicitly guided by some given body shape prior,  which is easy to obtain by existing pre-trained human parsing models \cite{humanparsing}.
In this way, both \textit{shape-related} and \textit{shape-erased} features are explicitly quantified while the discriminative nature of the two features can be independently maintained.

Specifically, we propose shape-erased feature learning paradigm that introduces orthogonality into representation to satisfy a relaxation of independent constraint. 
The representation is then decomposed into two sub-representations lying in two orthogonal subspaces for \textit{shape-related} and \textit{shape-erased} feature learning, respectively. 
By learning and covering most discriminative body shape feature in one subspace, the \textit{shape-erased} feature is forced to discover other modality-shared discriminative semantic concepts in the the other subspace as \textit{shape-related} feature is constrained in its orthogonal complement.
Under the above assumptions, we formulate this shape-erased feature learning paradigm from a mutual information perspective, and demonstrate that jointly learning  \textit{shape-erased} and \textit{shape-related} objectives achieves a conditional mutual information maximization between \textit{shape-erased} feature and identity discarding body shape information, thus enhancing the diversity of the learned representation explicitly. 
We finally design a Shape-Guided dIverse fEature Learning (SGIEL) framework that jointly optimizes \textit{shape-related} and \textit{shape-erased} objectives to learn modality-shared and discriminative integrated representation.
The contributions of our work are summarized as follows:
\begin{itemize}
    \item We propose a shape-erased feature learning  paradigm for VI-ReID that decorrelates  \textit{shape-erased} feature from \textit{shape-related} one by orthogonal decomposition. \textit{Shape-related} feature in one subspace is guided by body shape prior while \textit{shape-erased} feature is constrained in its orthogonal complement to discover more and other modality-shared discriminative semantic concepts, thus enhancing the diversity of the learned representation explicitly. 
    
    \item Based on the proposed shape-erased feature learning  paradigm, we design a Shape-Guided dIverse fEature Learning framework that jointly optimizes \textit{shape-related} and \textit{shape-erased} objectives to learn modality-shared and discriminative integrated representation.
    \item Extensive experiments on SYSU-MM01, RegDB, and HITSZ-VCM datasets demonstrate the effectiveness of our method.
\end{itemize}



\section{Related Work}
\label{sec:rw}
\subsection{Visible-Infrared Person Re-Identification}

To alleviate visible-infrared modality discrepancy and discover modality-shared discriminative features, researchers contributed numerous significant works in different levels of VI-ReID framework.
Specifically, in the feature learning level, Kansal \etal~\cite{sdl} designed a model to disentangle spectrum information and extract identity discriminative features to make cross-modal learning more efficient. Wu \etal~\cite{wu2020} exploited the same-modality similarity as a constraint to guide the learning of cross-modality similarity along with the alleviation of modality-specific information. Zhao \etal~\cite{c-ir-cons} designed a model which both learned the color-irrelevant features and aligned the identity-level feature distributions. Zhang \etal~\cite{fmcnet} proposed FMCNet to compensate the missing modality-specific information in the feature level.
In the input level, Ye \etal~\cite{caj} proposed a channel augmented joint learning strategy to improve the robustness against cross-modal variations. Wei \etal~\cite{Syncretic-Modality} introduced syncretic modality generative module to produce a new modality incorporating cross-modal features. 
In the model architecture level, most researchers implemented their frameworks in a one-stream~\cite{cm-nas} or two-stream~\cite{twostream,two-manner} manner, while some other researchers Li \etal~\cite{xmodal} claimed that an auxiliary modality was required as an assistant to bridge the huge gap between the two modalities.
Two existing related works consider a similar challenge including PAENet~\cite{paenet} and MPANet~\cite{nuances}. PAENet introduced eight attributes as annotations to learn the fine-grained semantic attribute information, which is a labeling-intensive journey in normal situations.
Wu \etal~\cite{nuances} proposed a joint Modality and Pattern Alignment Network to discover cross-modality nuances in different local patterns, however, they only add diverse constraints on the different activation maps and directly combine them together, which would fall into trivial solutions if one feature is enough to deal with the training set. 

Most existing VI-ReID approaches concentrated on reducing modality-specific features or converting them to modality-shared ones, without explicitly discovering diverse modality-shared features. Compared to previous works, we decorrelates \textit{shape-erased} feature from \textit{shape-related} one by orthogonal decomposition and by joint learning \textit{shape-erased} and \textit{shape-related} objectives, 
the diversity of the learned representation is explicitly  enhanced.


\begin{figure}[t]
  \centering
    \includegraphics[width=\linewidth]{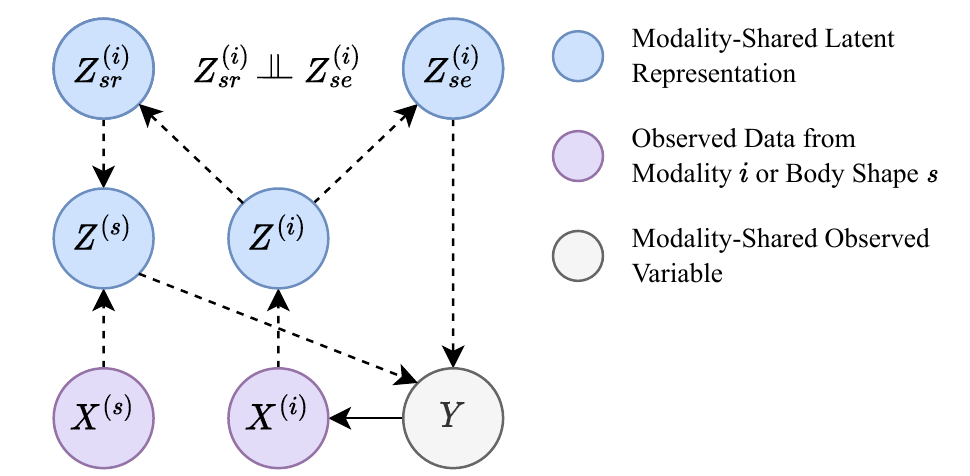}
  \caption{Graphical model of our method. The superscript $^{(s)}$ denotes body shape, and $^{(i)}$, where $i\!=\!1,2$, denotes different modalities (1(2) for visible(infrared)).   $Z_{sr}^{(i)}$ and $Z_{se}^{(i)}$ are assumed to be independent to explicitly quantify shape-related and shape-erased features.}
  \label{fig:graph}
  \vspace{-0.2cm}
\end{figure}
\vspace{-0.2cm}
\subsection{Semantic Parsing for Person Re-Identification}

Kalayeh \etal~\cite{semanticparsing} proposed the first work in person Re-ID community to adopt human semantic parsing to precisely localize arbitrary contours of various body parts. They claimed that compared to detection based method, semantic segmentation based method exhibited better pixel-level accuracy and capability of modeling arbitrary contours.
Song \etal~\cite{mask-guided} first introduced the binary segmentation masks to construct synthetic RGB-Mask pairs as inputs, and designed a mask-guided contrastive attention model to learn features separately from the body and background regions. 
Guo \etal~\cite{dualpart} applied a human parsing model to extract the binary human part masks and a self-attention mechanism to capture the soft latent (non-human) part masks. 
Zhu \etal~\cite{identity-guided} proposed the Identity-Guided Human Semantic Parsing approach to locate both the human body parts and personal belongings at pixel-level for person re-ID only with person identity labels. 
Hong \etal~\cite{clothchange} proposed a fine-grained shape-appearance mutual learning framework to learn fine-grained discriminative body shape knowledge to complement the cloth-unrelated knowledge in the appearance features.

For VI-ReID problem, one related work was contributed by Huang \etal~\cite{crossmodalmultitask}. They considered using person mask prediction as an auxiliary task with the help of a pre-trained human parsing model. In comparison to their work, we aim to discover richer modality-shared discriminative features in each subspace by erasing the body shape information to decorrelate shape-erased and shape-related features.


\section{Shape-Guided Diverse Feature Learning}
\label{sec:method}
In this section, we first present preliminaries of our method, and then describe our proposed shape-erased feature learning paradigm, and finally  introduce Shape-Guided dIverse fEature Learning (SGIEL) framework.

\subsection{Preliminary}

\paragraph{VI-ReID Setup.}Consider random variables $X^{(i)}$ and $Y$ representing data and label of VI-ReID, where $i\!=\!1$ for visible modality and $i\!=\!2$ for infrared modality. The observed values of  $X^{(i)}$ and $Y$ are used to build a dataset $D\! =\!\{D^{(i)}\}_{i=1}^2$, where $D^{(i)}\!=\!\{x_j^{(i)},y_j\}_{j=1}^{N_i}$. Samples of each modality are collected from the same group of $C$ persons, but the number of each identity's samples for each modality may arbitrary. Let $f$ and $g$ denote image encoder and classifier, the goal of VI-ReID is to learn an $f$ to extract representation $z^{(i)}=f(x^{(i)}) \in \mathbb{R}^n$ invariant to different modalities and different camera views. 

\paragraph{Body Shape Data.} We borrowed pre-trained Self-Correction Human Parsing (SCHP) model proposed in \cite{humanparsing} to segment body shape from background. 
Given a pixel of an image, we directly summed the probabilities of being a part of the head, torso, or limbs, predicted by SCHP, to create the body-shape map. Specifically, 
for each sample $x^{(i)}$ from dataset $D$, either visible or infrared, we used SCHP to produce its paired body shape map $x^{(s)}$ with the same image size and label, \ie, it is a one-to-one mapping between $D$ and its corresponding body shape data.  Let $f_{s}$ and $g_{s}$ denote body shape map encoder and classifier, the latent representation of $x^{(s)}$ is  $z^{(s)} =f_{s}(x^{(s)}) \in \mathbb{R}^m,$ $ m\!<\!n$.


\subsection{Shape-Erased Feature Learning Paradigm}
\label{sec:core}
In this section, we first explain the key independent assumption for explicitly quantifying \textit{shape-related} and \textit{shape-erased} features, and a relaxation to approximate it. Based on this relaxed independent constraint, we introduce the proposed Shape-Erased Feature Learning.


\begin{figure*}[t]
  \centering
  \includegraphics[width=\linewidth]{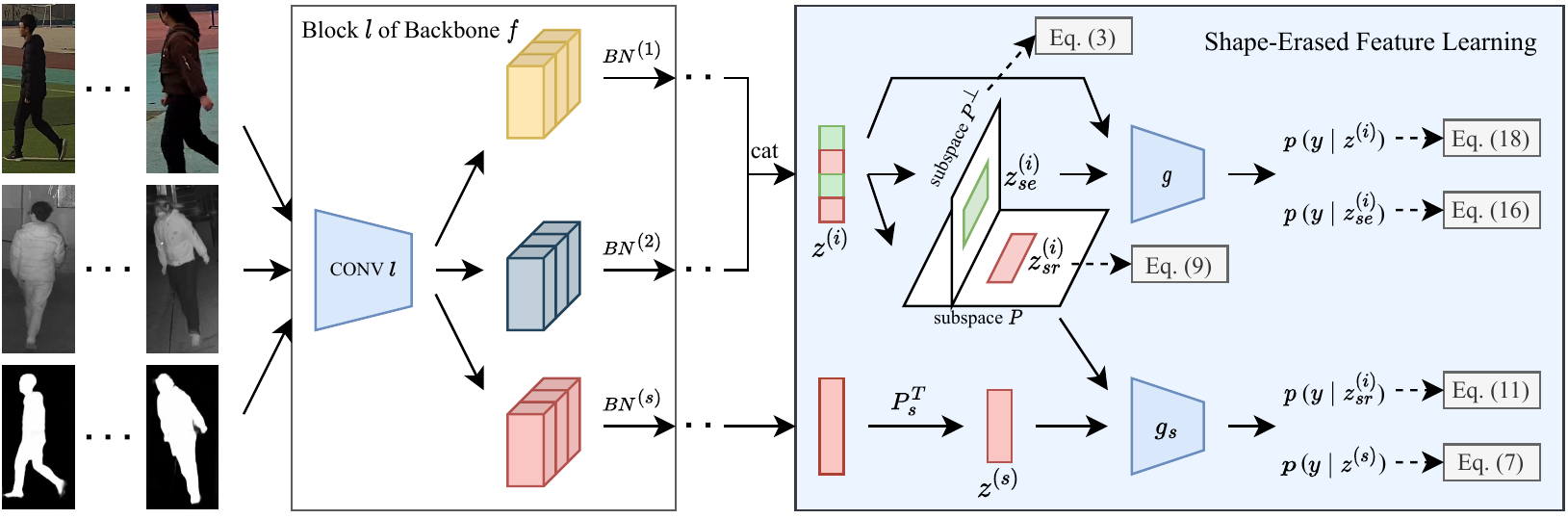}
  \caption{Shape-Guided Diverse Feature Learning. We utilize one shared backbone for visible, infrared images, and their body shape maps, while only BN layers are view-specific; ``cat'' refers to concatenating $z^{(i)}$ along batch dimension, where $i\!=\!1(2)$ for visible(infrared); In shape-erased feature learning paradigm, by regularizing $P$ to be semi-orthogonal (Eq.~\eqref{eq:ortho}), we decompose $z^{(i)}$ into \textit{shape-related} $z_{sr}^{(i)}$ and \textit{shape-erased} $z_{se}^{(i)}$.  $z_{sr}^{(i)}$ is learned to imitate and cover discriminative body shape features in subspace $P$ (Eq.~\eqref{eq:mse} and \eqref{eq:srce}), while $z_{se}^{(i)}$ is decorrelated to mine other modality-shared discriminative features in subspace $P^\perp$ (Eq.~\eqref{eq:sc}).}
  \label{fig:main}
  \vspace{-0.2cm}
\end{figure*}

\subsubsection{Independence between $Z_{sr}^{(i)}$ and $Z_{se}^{(i)}$}

We first formulate the main design of our shape-erased feature learning paradigm as a graphical model illustrated in Fig.~\ref{fig:graph}. It is assumed that modality-shared \textit{shape-related} feature, $Z_{sr}^{(i)}$, and modality-shared \textit{shape-erased} feature, $Z_{se}^{(i)}$, are independent from each other, and derived from an integrated representation $Z^{(i)}$ extracted in $X^{(i)}$, \ie, $Z^{(i)}\rightarrow Z_{sr}^{(i)},\ Z^{(i)}\rightarrow Z_{se}^{(i)}$. 
The independence between the two components, $Z_{sr}^{(i)} \perp \!\!\! \perp Z_{se}^{(i)}$, is necessary for learning any two features simultaneously without affecting each other. We formulate this independence as the following Eq.~\eqref{eq:partition},
\begin{equation}
    \begin{split}
        I(Z_{sr}^{(i)};Z_{se}^{(i)})&=0,
    \end{split}
\label{eq:partition}
\end{equation}
where  $I(\cdot;\cdot)$ denotes mutual information. As the mutual information estimation is complex and time-consuming, we relax the independence as an orthogonal constraint, and perform orthogonal decomposition to achieve the relaxed version of Eq. \eqref{eq:partition} as:
\begin{equation}
    \begin{split}
        z_{sr}^{(i)}&= P^Tz^{(i)}, \\
        z_{se}^{(i)}&=(I_n-PP^T) z^{(i)},
    \end{split}
\label{eq:orthodecompose}
\end{equation}
where $P \in \mathbb{R}^{n\times m} (m<n)$ denotes a semi-orthogonal matrix and $PP^T$ forms an orthogonal projector. In this way, \textit{shape-related} feature is learned in subspace $P$ while \textit{shape-erased} features is  learned in the orthogonal complement $P^\perp$, approximately satisfying the independent constraint. In practice, as $P$ is usually initialized by standard normal distribution, if $n \rightarrow \infty$, the probability that $P$ becomes a semi-orthogonal matrix goes to 1. 
To further enhance this orthogonality, we regularize $P$ by $L^1$-norm on the difference of each dimension between $P^TP$ and identity matrix $I_m$ by Eq. \eqref{eq:ortho}:
\begin{equation}
     \mathcal{L}_{ortho} = \frac{1}{m} \sum^{m}_{j=1} \| (P^TP)_j - (I_m)_j\|_1.
     \label{eq:ortho}
\end{equation}

\subsubsection{Shape-Erased Feature Learning}
As introduced in Section~\ref{sec:intro}, 
we aim to explicitly quantify $Z_{sr}^{(i)}$ and $Z_{se}^{(i)}$ so that $Z_{se}^{(i)}$ can infer identity $Y$  when discarding information used to describe $X^{(s)}$. This can be formulated as maximizing conditional mutual information between $Z_{se}^{(i)}$ and $Y$ given body shape $X^{(s)}$, \ie, $I(Z_{se}^{(i)};Y|X^{(s)})$:
\begin{equation}
     \max I(Z_{se}^{(i)};Y|X^{(s)})= I(Z_{se}^{(i)};Y) - I(Z_{se}^{(i)};Y;X^{(s)}),
\label{eq:sedef}
\end{equation}
where the first term represents mutual information between $Z_{se}^{(i)}$ and $Y$, and the second represents mutual information between $Z_{se}^{(i)}$, $Y$ and $X^{(s)}$. 

\paragraph{Maximize $I(Z_{se}^{(i)};Y)$.}

To optimize Eq.~\eqref{eq:sedef}, we can maximize the first term $ I(Z_{se}^{(i)};Y)$ by minimizing  cross-entropy ($l_{ce}(q,p)= -\sum_{k=1}^{C} p_k \log  q_k $) as Eq.~\eqref{eq:scid},
\begin{equation}
     \mathcal{L}_{seid} =  \mathbb{E}_{  (z_{se}^{(i)},y)\thicksim (Z_{se}^{(i)},Y)}   l_{ce}(g(z_{se}^{(i)}),y),\label{eq:scid}
\end{equation}
In the following, we will discuss how to estimate and minimize the second term $ I(Z_{se}^{(i)};Y;X^{(s)})$. 

\paragraph{Minimize $I(Z_{se}^{(i)};Y;X^{(s)})$.} Since $I(Y;X^{(s)})$ is intractable, we approximate it by the following two steps.

 \vspace{0.3cm}
 \noindent{(1) Approximate $I(Z_{se}^{(i)};Y;X^{(s)})$ by $I(Z_{se}^{(i)};Y;Z^{(s)})$} \\

Firstly, we consider a requirement that a representation $Z$ of $X$ can describe $Y$ at least as well as using the original data $X$ instead.
This requirement is known as \textit{sufficiency}~\cite{suff} that can be defined as follows:
\\

\noindent{Definition 1 (Sufficiency). \textit{A representation $Z$ of $X$ is sufficient for $Y$ if and only if:}}
\begin{equation}
 I(X; Y|Z) = 0 \iff     I(X;Y)=I(Z;Y).
    \label{eq:suff}
\end{equation}
For $Z^{(s)}$, if the classification loss $\mathcal{L}_{sid}$ is minimized,
\begin{equation}
     \mathcal{L}_{sid} =  \mathbb{E}_{  (z^{(s)},y)\thicksim (Z^{(s)},Y)}   l_{ce}(g_s(z^{(s)}),y),\label{eq:sid}
\end{equation}
then following \cite{infodropout}, we can assume $Z^{(s)}$ of $X^{(s)}$ for $Y$ is \textit{sufficient}, and thus we replace $I(Z_{se}^{(i)};Y;X^{(s)})$ with $I(Z_{se}^{(i)};Y;Z^{(s)})$ in Eq.~\eqref{eq:sedef} combining Eq.~\eqref{eq:suff}. This replacement can be formulated as the following \textbf{Theorem 1}.

\begin{theorem}
If representation $Z^{(s)}$ of $X^{(s)}$ is \textbf{sufficient} for $Y$, then $I(Z^{(i)}_{se};Y;X^{(s)})=I(Z^{(i)}_{se};Y;Z^{(s)})$.
\end{theorem}
The proof can be found in supplementary material.

\vspace{0.3cm}
\noindent {(2) Approximate $I(Z_{se}^{(i)};Y;Z^{(s)})$ by $I(Z_{se}^{(i)};Y;Z_{sr}^{(i)})$}\\

Secondly, we hope shape-related feature $Z_{sr}^{(i)}$ can fully represent real body shape feature $Z^{(s)}$, so that if $Z_{sr}^{(i)}\equiv Z^{(s)}$, then $I(Z_{se}^{(i)};Y;Z^{(s)})$ can also be approximated by $I(Z_{se}^{(i)};Y;Z_{sr}^{(i)})$: 
\begin{equation}
\begin{split}
    I(Z_{se}^{(i)};Y;X^{(s)})&=I(Z_{se}^{(i)};Y;Z^{(s)})\\
    &=I(Z_{se}^{(i)};Y;Z_{sr}^{(i)})\le I(Z_{se}^{(i)};Z_{sr}^{(i)})=0,
\end{split}
    \label{eq:upbound}
\end{equation} 
which is upper-bounded by Eq.~\eqref{eq:partition}. To achieve a $Z_{sr}^{(i)}$  fully representing $Z^{(s)}$, 
as there exists a one-to-one mapping between $Z_{sr}^{(i)}$ and $Z^{(s)}$, we maximize $I(Z_{sr}^{(i)};Z^{(s)})$ by minimizing element-wise mean squared error (MSE) to guide $Z_{sr}^{(i)}$ to imitate $Z^{(s)}$ as Eq.~\eqref{eq:mse},
\begin{equation}
     \mathcal{L}_{srmse} =\mathbb{E}_{ (z_{sr}^{(i)},z^{(s)})\thicksim (Z_{sr}^{(i)},Z^{(s)}) } \frac{\| z_{sr}^{(i)} - z^{(s)}\|_2^2}{m} ,\label{eq:mse}
\end{equation}
where $\|\cdot \|_2$ denotes $l^2$-norm. 

Moreover, to 
reduce cross-view discrepancy between $Z_{sr}^{(i)}$ of $X^{(i)}$ and $Z^{(s)}$ of $X^{(s)}$, we aim to minimize the following conditional mutual information $I(X^{(i)};Z_{sr}^{(i)} | X^{(s)}) $:
\begin{equation}
    \min I(X^{(i)};Z_{sr}^{(i)} | X^{(s)}),\label{eq:crossviewcmi}
\end{equation}
denoting the remaining information in $Z_{sr}^{(i)}$ given the view of $X^{(s)}$.
To minimize Eq.~\eqref{eq:crossviewcmi}, we follow \cite{mireid} to approximate an upper bound of it as a Kullback–Leibler (KL) divergence between $p(y|z_{sr}^{(i)})$ and $p(y|z^{(s)})$. The proof of that can be found in supplementary material. For simplicity, we directly minimize cross-entropy loss as Eq.~\eqref{eq:srce} for the remaining information entropy term in KL divergence only depending on target distribution $p(y|z^{(s)})$,
\begin{equation}
     \mathcal{L}_{srkl} = \mathbb{E}_{ (z_{sr}^{(i)},z^{(s)})\thicksim (Z_{sr}^{(i)},Z^{(s)}) } l_{ce}( g_s(z_{sr}^{(i)}),g_s(z^{(s)})).\label{eq:srce}
\end{equation}
Combining Eq.~\eqref{eq:mse} and \eqref{eq:srce}, the final shape-related objective becomes:
\begin{equation}
     \mathcal{L}_{sr} = \mathcal{L}_{srmse} + \mathcal{L}_{srkl}.\label{eq:sr}
\end{equation}

Minimizing Eq.~\eqref{eq:sr}, we can represent $Z^{(s)}$ by $Z_{sr}^{(i)}$ approximately. If both \textit{sufficiency} of $Z^{(s)}$ is achieved and Eq.~\eqref{eq:sr} is minimized, then,
\begin{equation}
    I(Z_{se}^{(i)};Y|X^{(s)})\ge I(Z_{se}^{(i)};Y)
\end{equation}
will hold by Eq.~\eqref{eq:upbound}. In this way, \textit{shape-erased} features can be learned by minimizing classification loss in Eq.~\eqref{eq:scid} as information used to describe discriminative body shape feature are approximately discarded by orthogonal decomposition in Eq.~\eqref{eq:orthodecompose}.

\paragraph{Eliminate Modality-Specific Information.}
It is to be noted that both $Z_{sr}^{(i)}$ and $Z_{se}^{(i)}$ are assumed to be shared features between the two modalities. $Z_{sr}^{(i)}$ is learned to imitate body shape representation $Z^{(s)}$ to be modality-shared naturally; For $Z_{se}^{(i)}$, we eliminate modality-specific  information in a mutual manner as follows:
\begin{equation}
    \min I(X^{(1)};Z_{se}^{(1)}|X^{(2)})+I(X^{(2)};Z_{se}^{(2)}|X^{(1)}).
\end{equation}
Similar to Eq.~\eqref{eq:srce}, we approximated it as a cross-modal cross-entropy between $p(y|z_{se}^{(1)})$ and $p(y|z_{se}^{(2)})$ and vice versa:
\begin{equation}
     \mathcal{L}_{sekl} = \mathbb{E}_{ (z_{se}^{(i)},z_{se}^{(3-i)})\thicksim (Z_{se}^{(i)},Z_{se}^{(3-i)}) }  l_{ce}(g(z_{se}^{(i)}),g(z_{se}^{(3-i)})),\label{eq:scce}
\end{equation}
where $i=1,2$.

Combining Eq.~\eqref{eq:scid} and \eqref{eq:scce}, the shape-erased objective can be formulated as:
\begin{equation}
     \mathcal{L}_{se} = \mathcal{L}_{seid} + \mathcal{L}_{sekl}.\label{eq:sc}
\end{equation}

\subsection{Overall Framework}
\label{sec:fw}
In Section \ref{sec:core}, we decompose representation $z^{(i)}$ into two orthogonal components named \textit{shape-related} $z_{sr}^{(i)}$ and \textit{shape-erased} $z_{se}^{(i)}$. To further enhance the discriminative and modality-shared natures of $z^{(i)}$,  we 
apply commonly used classification loss $\mathcal{L}_{id}$ and triplet loss~\cite{triplet} $\mathcal{L}_{triplet}$ on $z^{(i)}$. 
For triplet pairs, we find the hardest positive and negative pairs among all samples in a mini-batch, consisting of both visible and infrared samples.

Similar to $\mathcal{L}_{sekl}$, we apply the following Eq.~\eqref{eq:gkl} for eliminating cross-modal discrepancy in a mutual way:
\begin{equation}
     \mathcal{L}_{kl} = \mathbb{E}_{ (z^{(i)},z^{(3-i)})\thicksim (Z^{(i)},Z^{(3-i)}) }  l_{ce}(g(z^{(i)}),g(z^{(3-i)})).\label{eq:gkl}
\end{equation}

Combining $\mathcal{L}_{id}$, $\mathcal{L}_{triplet}$ and \eqref{eq:gkl}, the integrated representation objective can be formulated as:
\begin{equation}
     \mathcal{L}_{int} =\mathcal{L}_{id}+\mathcal{L}_{triplet}+\mathcal{L}_{kl} .\label{eq:g}
\end{equation}

Moreover, we implement a re-weighting mechanism to focus on more difficult objective between $\mathcal{L}_{sr}$ and $\mathcal{L}_{se}$ during the whole training process. Let $\theta_t$ denote parameters to be optimized at training iteration $t$-th, we measure this difficulties by comparing the norms of $\frac{\partial \mathcal{L}_{sr}(\theta_t) }{\partial \theta_t}$ and $\frac{\partial \mathcal{L}_{se}(\theta_t) }{\partial \theta_t}$. The objective with a larger gradient norm is regarded to be more difficult  at iteration $t$-th.
To save computation cost, following~\cite{featgrad,featgrad2}, we approximate the actual parameter-level gradients by the representation-level gradients, \ie, replacing $\frac{\partial \mathcal{L}(\theta_t) }{\partial \theta_t}$ with $\frac{\partial \mathcal{L}(\theta_t) }{\partial z^{(i)}}$. Our final re-weighting mechanism is as follows:
\begin{equation}
\small
    \begin{split}
        \alpha_t^{sr} &= {\|\frac{\partial \mathcal{L}_{sr}(\theta_t) }{\partial  z^{(i)}}\|_2^2}/({\|\frac{\partial \mathcal{L}_{sr}(\theta_t) }{\partial  z^{(i)}}\|_2^2+\|\frac{\partial \mathcal{L}_{se}(\theta_t) }{\partial  z^{(i)}}\|_2^2}) , \\
        \alpha_t^{se} &= {\|\frac{\partial \mathcal{L}_{se}(\theta_t) }{\partial  z^{(i)}}\|_2^2}/({\|\frac{\partial \mathcal{L}_{sr}(\theta_t) }{\partial  z^{(i)}}\|_2^2+\|\frac{\partial \mathcal{L}_{se}(\theta_t) }{\partial  z^{(i)}}\|_2^2}) ,
    \end{split} \label{eq:weightingfactor}
\end{equation}

The overall training loss can be summarized as the following Eq.~\eqref{eq:overall} :
\begin{equation}
    \mathcal{L}_{train}= \mathcal{L}_{int} + \alpha_t^{sr} \mathcal{L}_{sr} + \alpha_t^{se} \mathcal{L}_{se} + \mathcal{L}_{ortho} + \mathcal{L}_{sid}.\label{eq:overall}
\end{equation}
The overall framework of our method is illustrated in Fig. \ref{fig:main}. To reduce the computation and GPU memory consumption, one modeling backbone is shared for three types of data (two modalities and body shape map). Considering the distribution gap among them, similar to AdaBN \cite{adabn}, for each Batch Normalization (BN) layer in the backbone, we implement three new parameter-specific BNs to replace it as a normalization across different distributions. Following the design of BNNeck by \cite{bnneck}, we perform three     parameter-specific BNNecks after the backbone.

\begin{table*}[htbp]
  \centering
 \small
       \caption{Comparison with state-of-the-art methods in SYSU-MM01 under single-shot setting. The performance is shown by Rank-k accuracy (\%) and mAP (\%). The best results and the second are \textbf{bold} and \underline{underlined} marked, respectively. ``C'' refers to concatenated the output representations of two checkpoints of our model. ``2x'' refers to 2 times of size of parameters contained in ResNet50.}
\vspace{-0.1cm}
       
    \begin{tabular}{c|c|c|cccc|cccc}
    \hline
    \multirow{2}*{Params} & \multirow{2}*{Method} & \multirow{2}*{Venue} & \multicolumn{4}{c|}{All Search} & \multicolumn{4}{c}{Indoor Search} \\
\cline{4-11}          &       &       & Rank-1 &  Rank-10  & Rank-20 & mAP   & Rank-1 &  Rank-10  & Rank-20 & mAP \\\hline
    $\thickapprox$1x   & CM-NAS~\cite{cm-nas} & CVPR'21 & 61.99 & 92.87 & 97.25 & 60.02 & 67.01 & 97.02 & 99.32 & 72.95 \\
    $\thickapprox$1x   & CAJL~\cite{caj}  & ICCV'21 & 69.88 & 95.71 & 98.46 & 66.89 & 76.30  & 97.90  & 99.50  & 80.40 \\
    $\thickapprox$1x   & MPANet~\cite{nuances}  & CVPR'21 & 70.58 & 96.21 & 98.8  & 68.24 & 76.64 & 98.21 & 99.57 & 80.95 \\
    $\thickapprox$1x   & MMN~\cite{mmn}   & ACMMM'21 & 70.60  & 96.20  & 99.00    & 66.90  & 76.20  & 97.20  & 99.30  & 79.60 \\ \hline
    $\thickapprox$1.75x & MTL~\cite{crossmodalmultitask}   & PR'22 & 67.25 & 95.38 & 98.46 & 64.29 & 69.58 & 96.66 & 99.03 & 74.37 \\
    $\thickapprox$1.25x & PAENet~\cite{paenet} & ACMMM'22 & 74.22 & \textbf{99.03} & \textbf{99.97} & \textbf{73.90}  & 78.04 & \textbf{99.58} & \textbf{100.00} & \textbf{83.54} \\
    $\thickapprox$2x   & MSCLNet~\cite{msclnet} & ECCV'22 & \underline{76.99} & \underline{97.93} & \underline{99.18} & 71.64 & \underline{78.49} & \underline{99.32} & \underline{99.91} & 81.17 \\
    \hline
    $\thickapprox$1x   & Ours  & -     & 75.18 & 96.87 & 99.13 & 70.12 & 78.40  & 97.46 & 98.91 & 81.20 \\
    $\thickapprox$2x  & Ours (C)  & -     & \textbf{77.12} & 97.03 & 99.08 & \underline{72.33} & \textbf{82.07} & 97.42 & 98.87 & \underline{82.95} \\\hline
    \end{tabular}%
  \label{tab:sysumm01}%
\vspace{-0.1cm}
  
\end{table*}%
\begin{table*}[htbp]
  \centering
   \small
    \caption{Comparisons of our method with state-of-the-art methods on HITSZ-VCM dataset. Rank-k accuracy (\%) and mAP (\%) are reported. The best results and the second are in \textbf{bold} and \underline{underlined}, respectively.}
\vspace{-0.1cm}

    \begin{tabular}{c|c|c|cccc|cccc}
    \hline
    \multirow{2}*{Strategy} & \multirow{2}*{Method} & \multirow{2}*{Venue} & \multicolumn{4}{c|}{Infrared to Visible} & \multicolumn{4}{c}{Visible to Infrared} \\
\cline{4-11}          &       &       & Rank-1 & Rank-5 & Rank-10 & mAP   & Rank-1 & Rank-5 & Rank-10 & mAP \\
    \hline
    Video & MITML \cite{vcm} & CVPR'22 & \underline{63.74} & \underline{76.88} & \underline{81.72} & 45.31 & 64.54 & \underline{78.96} & 82.98 & 47.69 \\
    \hline
    \multirow{7}*{Image} & LbA \cite{lba}   & ICCV'21 & 46.38 & 65.29 & 72.23 & 30.69 & 49.30  & 69.27 & 75.90  & 32.38 \\
          & MPANet \cite{nuances} & CVPR'21 & 46.51 & 63.07 & 70.51 & 35.26 & 50.32 & 67.31 & 73.56 & 37.80 \\
          & DDAG \cite{ddag} & ECCV'20 & 54.62 & 69.79 & 76.05 & 39.26 & 59.03 & 74.64 & 79.53 & 41.50 \\
          & VSD \cite{mireid}  & CVPR'21 & 54.53 & 70.01 & 76.28 & 41.18 & 57.52 & 73.66 & 79.38 & 43.45 \\
          & CAJL \cite{caj} & ICCV'21 & 56.59 & 73.49 & 79.52 & 41.49 & 60.13 & 74.62 & 79.86 & 42.81 \\
\cline{2-11}          & Baseline & -     & 62.02 & 75.35 & 81.35 & \underline{47.05} & \underline{64.90}  & 78.64 & \underline{83.68} & \underline{48.21} \\
          & Ours  & -     & \textbf{67.65}&	\textbf{80.32}&	\textbf{84.73}&	\textbf{52.30}&	\textbf{70.23}&	\textbf{82.19}&	\textbf{86.11}&	\textbf{52.54} \\
    \hline
    \end{tabular}%
  \label{tab:vcm}%
\vspace{-0.2cm}
  
\end{table*}%
\begin{table}[htbp]
  \centering
  \small
    \caption{Comparison with state-of-the-art methods in RegDB. The performance is shown by Rank-1 (\%) and mAP (\%). The best results and the second are \textbf{bold} and \underline{underlined} marked, respectively.}
    \begin{tabular}{c|cc|cc}
    \hline
    \multirow{2}*{Method} & \multicolumn{2}{c|}{Infrared to Visible} & \multicolumn{2}{c}{Visible to Infrared} \\
\cline{2-5}          & Rank-1 & mAP   & Rank-1 & mAP \\
    \hline
    CM-NAS \cite{cm-nas} & 82.57 & 78.31 & 84.54 & 80.32 \\
    CAJL \cite{caj} & 84.75 & 77.82 & 85.03 & 79.14 \\
    MPANet \cite{nuances} & 82.8  & 80.7  & 83.7  & 80.9 \\
    MMN \cite{mmn}  & 87.5  & 80.5  & 91.6  & 84.1 \\
    \hline
    MTL \cite{crossmodalmultitask}  & 88.34 & 84.06 & 89.91 & 85.64 \\
    PAENet \cite{paenet}& \textbf{95.35} & \textbf{89.98} & \textbf{97.57} & \textbf{91.41} \\
    MSCLNet\cite{msclnet}  & 83.86 & 78.31 & 84.17 & 80.09 \\
    \hline
    Ours &\underline{91.07}	&\underline{85.23}&	\underline{92.18}	&\underline{86.59} \\
    \hline
    \end{tabular}%
  \label{tab:regdb}%
\vspace{-0.3cm}
  
\end{table}%
\section{Experiments}
\label{sec:exp}
To validate the effectiveness of our method, we conducted experiments on two benchmark datasets, SYSU-MM01~\cite{sysumm01} and RegDB~\cite{regdb}. We also evaluated our method on a large-scale video-based VI-ReID dataset named HITSZ-VCM~\cite{vcm}.
Furthermore, we performed ablation study to validate the effectiveness of each component in our method. 

\subsection{Dataset and Evaluation Protocol}
SYSU-MM01~\cite{sysumm01} contains four(two) cameras for capturing visible(infrared) images. The query and gallery set are collected from another 96 identities containing 3,803 infrared images and 301 randomly sampled visible images under single-shot setting.
RegDB~\cite{regdb} is captrured by one pair of visible and thermal cameras. It is collected from 412 identities. The identities are randomly and equally split into a training set and a testing set. RegDB has two testing protocols: infrared to visible, which means infrared serves as query, and visible to infrared, meaning visible as query. 

HITSZ-VCM~\cite{vcm}  is the first large-scale video-based VI-ReID dataset captured by  6  visible and 6 infrared cameras. There are 251,452 visible images and 211,807 infrared images. Every 24 consecutive images are regarded as a tracklet. 927 identities are divided into 500 for training and 427 for testing. The testing set contains 5,159 infrared and 5,643 visible tracklets. The testing protocol is similar to RegDB's but in a \textit{tracklet-to-tracklet} way.

To quantitatively evaluate the performance of our proposed model, Cumulative Matching Characteristic curve (CMC) and mean Average Precision(mAP) are adopted as the evaluation metrics on all three datasets.

\subsection{Implementation Details}
Following \cite{caj,agw}, we adopted an ImageNet pre-trained ResNet50~\cite{resnet} as our backbone, and replaced the average pooling layer with GEM-pooling similar to \cite{agw}. We adopted Random Channel Exchangeable Augmentation and Channel-Level Random Erasing proposed in \cite{caj} and exponential moving average (EMA) model, similar to~\cite{mmt}, for our method and baseline. Only the EMA model is used for testing. We replaced $z^{(s)}$ in Eq.~\eqref{eq:srce} with a temporal ensemble $z^{(s)}$ produced by EMA model. The dimension $m(n)$ of $P$ and $P_s$ is set to $512(2048)$.
The three sets of BNs were initialized to be identical to each other. 

We used SGD optimizer with initial learning rate of 0.1 for randomly initialized parameters and 0.01 for pre-trained parameters. 
For SYSU-MM01 and RegDB dataset, we trained the model 100 epochs and decreased the learning rate by a factor of 10 at the 20-th and 50-th epoch. We randomly sampled 8 identities, each of which involved 4 visible and 4 infrared images to build a mini-batch.
For HITSZ-VCM dataset, we trained the model 200 epochs and decreased the learning rate by a factor of 10 at the 35-th and 80-th epoch. We randomly sampled 8 identities, each of which involved 2 visible and 2 infrared tacklets, and randomly sampled 3 frames per tracklet to build a mini-batch.

\subsection{Comparison with State-of-the-Art Methods}
\paragraph{Comparison on SYSU-MM01 and RegDB.} We compared our method with existing state-of-the-art methods for VI-ReID, including CM-NAS~\cite{cm-nas}, CAJL~\cite{caj}, MPANet~\cite{nuances}, MMN~\cite{mmn}, MTL~\cite{crossmodalmultitask}, PAENet~\cite{paenet}, MSCLNet~\cite{msclnet}. For SYSU-MM01 in Table~\ref{tab:sysumm01}, ``Params'' indicates which times of ResNet50 parameter size the backbone is. To achieve a comparable size with ``$\thickapprox$2x params'' methods, we directly concatenated the output representations of two checkpoints of our model.  Our method outperformed most models with comparable size of parameters according to  Rank-1 and mAP metrics. Note that  PAENet introduced eight extra fine-grained attributes as annotations for each sample. For RegDB in Table~\ref{tab:regdb}, our method also achieved comparable results with previous methods.

\paragraph{Comparison on HITSZ-VCM.} There are two testing strategies named video-based and image-based. Following \cite{vcm}, we conducted an average pooling layer on frame-level representations to form a tracklet-level representation during testing. The results are shown in Table~\ref{tab:vcm}. ``Baseline'' refers to only optimizing with one classification and triplet loss. Our method outperformed all existing methods, and compared to the baseline, our method achieved 5\% improvements on most metrics.

\subsection{Ablation Study and Analysis}
In this section, we conducted ablation study to evaluate the contribution of each component in our proposed Shape-Guided dIverse fEature Learning (SGIEL). All experiments were performed on SYSU-MM01 under all search, single-shot setting using the same baseline. We tuned the hyper-parameters of each experiment carefully. The results are summarized in Table~\ref{tab:ablation} and \ref{tab:ortho} and described as follows.
\vspace{-0.2cm}
\paragraph{Effectiveness of Erasing Body Shape.}
We first introduce the baseline (Exp 1) of SGIEL, which using the same backbone but only optimized with one classification and triplet loss on representation $z^{(i)}$. In Exp 2, we add $\mathcal{L}_{gkl}$ to form the objective Eq.~\eqref{eq:g}.

To evaluate the effectiveness of erasing body shape information, in Exp 3, we only applied \textit{shape-related} objective, $\mathcal{L}_{sr}$ in Eq.~\eqref{eq:sr}, which indicated the model was forced to only concentrate on \textit{shape-related} feature without considering others, resulting in degeneration of performance compared to Exp 2. However, when applying shape erased feature learning paradigm with $\mathcal{L}_{se}$ in Eq.~\eqref{eq:sc} (Exp 4), we achieved 3\% in Rank-1 and 4\% in mAP improvement compared to Exp 3. Moreover, in Exp 5, adding an additional orthogonal constraint ($\mathcal{L}_{ortho}$) to strengthen the orthogonality of \textit{shape-related} and \textit{shape-erased} features also boosted the performance. The above ablation study demonstrated the effectiveness of erasing body shape in our SGIEL. 

\begin{table}[tbp]
  \centering
    \caption{The individual improvements of our method performed on SYSU-MM01 with all search, single shot setting. Compared to Exp 2, the baseline (Exp 1) only contains one classification and triplet loss. ``SEFEL'' denotes our Shape-Erased FEature Learning paradigm. ``'$\alpha$'' denotes our re-weighting mechanism.}
      \scalebox{0.89}{
    \begin{tabular}{c|c|ccc|c|cc}
    \hline
    \multirow{3}*{Exp} & \multicolumn{5}{c|}{Component}                     & \multirow{3}*{Rank-1} & \multirow{3}*{mAP} \\
\cline{2-6}           & \multirow{2}*{$\mathcal{L}_{kl}$} & \multicolumn{3}{c|}{SEFEL} & \multirow{2}*{$\alpha$} &       &  \\
\cline{3-5}            &       & $\mathcal{L}_{ortho}$ & $\mathcal{L}_{sr}$  & $\mathcal{L}_{se}$   &       &       &  \\
    \hline
    1(base) &     &       &       &       &       &     68.55  & 65.32 \\\hline
    2          & \checkmark &       &       &       &       &   70.73    & 66.44 \\\hline
    3     &  \checkmark &       & \checkmark &       &       & 69.60 & 64.06 \\\hline
    4     & \checkmark &       & \checkmark & \checkmark &       & 72.00    & 68.57 \\\hline
    5     & \checkmark & \checkmark & \checkmark & \checkmark &       & 73.52 & 69.05 \\\hline
    6(full) & \checkmark & \checkmark & \checkmark & \checkmark & \checkmark &  75.18 & 70.12 \\\hline
    \end{tabular}%
    }
  \label{tab:ablation}%
\end{table}%

\begin{table}[tbp]
  \centering
  \caption{Variants of orthogonal constraint. ``Num of Proj'' denotes the number of linear projectors. }
    \begin{tabular}{cc|cc}
    \hline
    Num of Proj & $\mathcal{L}_{ortho}$ & Rank-1 & mAP \\
    \hline
    2     &      & 70.18 & 66.36 \\
    2     & \checkmark     & 72.76 & 68.53 \\
    1     &      &   73.49    & 68.97 \\ \hline
    1     & \checkmark     & 75.18 & 70.12 \\
    \hline
    \end{tabular}%
  \label{tab:ortho}%
\vspace{-0.2cm}
  
\end{table}%
\vspace{-0.2cm}

\paragraph{Effectiveness of Orthogonal Constraint.}
In Section \ref{sec:core}, we demonstrated the necessity of independent constraint on \textit{shape-related} and \textit{shape-erased} features, and introduced orthogonal decomposition on representation as a relaxation of independent constraint. We evaluate this necessity and our design on orthogonal constraint experimentally in Table~\ref{tab:ortho}. 
\vspace{-0.2cm}

\paragraph{- Necessity of Orthogonal Constraint.}

To evaluate the necessity of orthogonal constraint, we designed a fully unconstrained model as follows. We only replaced orthogonal projector $P$ in our method with two unconstrained projectors $P_1 \in \mathbb{R}^{n\times m_1},P_2 \in \mathbb{R}^{n\times m_2}$ independently learned by $\mathcal{L}_{sr}$ and $\mathcal{L}_{se}$, respectively, (\ie, $z_{sr}^{(i)}=P_1^Tz^{(i)}, \ z_{se}^{(i)}=P_2^Tz^{(i)}$). We reported this experiment as the 1-st line in Table~\ref{tab:ortho}, which paid a huge loss of performance. 

To further evaluate this necessity, similar to Exp 4 in Table~\ref{tab:ablation}, we directly discarded $\mathcal{L}_{ortho}$ in our method (but kept the decomposition process), and also resulted in a performance degeneration. We reported this experiment as the 3-rd line in Table~\ref{tab:ortho}. We observed across datasets that cosine similarity between individual dimensions in $P$ remains about 0.015 in average after initialization, and increased to 0.03 in average when convergence, which would be harmful to discover body-shape-unrelated features.

\paragraph{- Design of Orthogonal Constraint.} To evaluate the design of orthogonal constraint in shape erased feature learning, based on the above mentioned fully unconstrained model, we added an orthogonal constraint loss similar to $\mathcal{L}_{ortho}$ but between the two projectors, $P_1$ and $P_2$. We carefully tuned the hyper-parameters including $m_1=512$ and $m_2=1024$, and reported the final results as the 2-nd line in Table~\ref{tab:ortho}, achieving a lower performance compared to ours.

\begin{figure}[t]
  \centering
    \includegraphics[width=\linewidth]{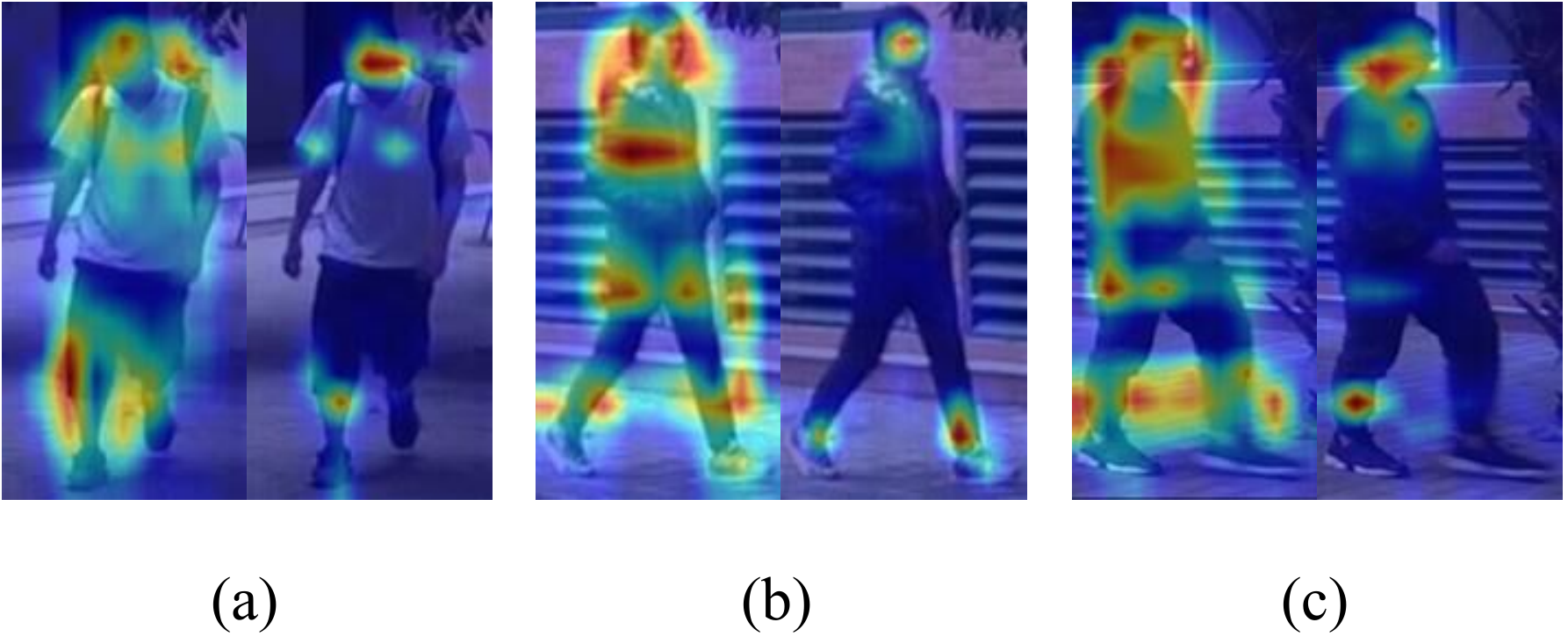}
  \caption{Visualization of \textit{shape-related} objective $\mathcal{L}_{sr}$ and \textit{shape-erased} objective $\mathcal{L}_{se}$ separately on the last feature map of the backbone through Grad-CAM++. For each pair of images, the left/right images visualized the gradient CAM produced by $\mathcal{L}_{sr}$/$\mathcal{L}_{se}$.}
  \label{fig:visual}
\vspace{-0.2cm}
  
\end{figure}

\paragraph{Visualization on Feature Maps.} We visualized \textit{shape-related} objective $\mathcal{L}_{sr}$ and \textit{shape-erased} objective $\mathcal{L}_{se}$ separately on the last feature map of the backbone through Grad-CAM++~\cite{gradcam}. As illustrated in Fig.~\ref{fig:visual}, On the left of each pair of  gradient CAM images produced by $\mathcal{L}_{sr}$, the activation mostly focused on the contour of their figures, while the gradient CAM of the right one produced by $\mathcal{L}_{se}$ is more centralized in complementary parts, such as head and legs.

\section{Conclusion}
In this work, we investigate how to learn diverse features in VI-ReID and try to erase body-shape-related semantic concepts in the learned features to force the ReID model to extract more and other modality-shared discriminative features.
We propose shape-erased feature learning paradigm that decorrelates features in two orthogonal subspaces of shape-related and shape-erased ones. By learning discriminative body shape feature in one subspace, the shape-erased feature is forced to discover other modality-shared discriminative semantic concepts in the the other subspace as shape-related feature is constrained in its orthogonal complement. We demonstrate that jointly learning the two objectives achieves a conditional mutual information maximization between shape-erased feature and identity discarding body shape information, thus enhancing the diversity of the learned representation explicitly. 
Extensive experiments on SYSU-MM01, RegDB, and HITSZ-VCM datasets demonstrate the effectiveness of our method.

\noindent \textbf{Acknowledgements} This work was supported partially by the National Science Foundation for Young Scientists of China (62106288) and the Guangdong Basic and Applied Basic Research Foundation (2023A1515012974). We would like to express our sincere thanks to Wenhang Ge from the Hong Kong University of Science and Technology (Guangzhou), for his valuable discussions. 
{\small
\bibliographystyle{ieee_fullname}
\bibliography{egbib}
}

\begin{onecolumn}

\appendix
\begin{center}
    \Large
    \textbf{Supplementary Material for Shape-Erased Feature Learning for Visible-Infrared Person Re-Identification}
    \\[10pt]
\end{center}
\renewcommand\thetable{S\arabic{table}}

\setcounter{equation}{0}
\setcounter{theorem}{0}
\setcounter{table}{0}

\vspace{10px}

\section{On Minimizing  $I(Z_{se}^{(i)};Y;X^{(s)})$}
\label{A}
\renewcommand{\theequation}{\Alph{section}.\arabic{equation}}

In Appendix \ref{A}, we will proof $I(Z_{se}^{(i)};Y;X^{(s)})$ can be approximately upper-bounded by 0, so that $\max I(Z_{se}^{(i)};Y|X^{(s)})$ can be lower-bounded by $\max I(Z_{se}^{(i)};Y)$ (Eq. (4), (13) in the paper). We first enumerate the following four hypotheses for minimizing  $I(Z_{se}^{(i)};Y;X^{(s)})$.
\vspace{0.3cm}

\noindent{\textit{Hypothesis:}}
\begin{enumerate}
    \item Following \cite{suppppp}, if $Z^{(s)}$ is a representation of $X^{(s)}$, then we state that $Z^{(s)}$ is conditionally independent from any other variable in the system once $X^{(s)}$ is observed (\eg $Z^{(s)}$ can be a deterministic function of $X^{(s)}$): $$\forall A,B, \quad I(A;Z^{(s)}|X^{(s)},B)\!=\!0.$$
    \item  (Eq. (6) in the paper) $Z^{(s)}$ is a \textit{sufficient} representation of $X^{(s)}$ for $Y$, \ie, $I(Y;X^{(s)}|Z^{(s)})\!=\!0$.
    \item (Eq. (8) in the paper) $Z_{sr}^{(i)}$ can fully represent $Z^{(s)}$, \ie, $Z^{(s)}\equiv Z_{sr}^{(i)}$.
    \item  (Eq. (1), (2) in the paper) The orthogonality between $z^{(i)}_{se}$ and $z_{sr}^{(i)}$ can be regarded as a relaxation of independence, \ie, $$\forall \ (z_{sr}^{(i)},z_{se}^{(i)}) \thicksim (Z_{sr}^{(i)},Z_{se}^{(i)}),\ z_{sr}^{(i)}\perp z_{se}^{(i)} \Longrightarrow I(Z_{sr}^{(i)};Z_{se}^{(i)})\thickapprox 0.  $$
\end{enumerate}
\textit{Hypothesis} 2 can be satisfied by \textbf{Proposition 1} in Appendix~\ref{B}. To approximate \textit{Hypothesis} 3, we minimize element-wise mean squared error (MSE) between them (Eq. (9) in the paper), and it is to be noted that the gradient of $z^{(s)}$ is discarded.
\vspace{0.2cm}
\begin{theorem}
If representation $Z^{(s)}$ of $X^{(s)}$ is \textbf{sufficient} for $Y$, then $I(Z^{(i)}_{se};Y;X^{(s)})=I(Z^{(i)}_{se};Y;Z^{(s)})$.
\begin{proof}
Obviously, $I(Z^{(i)}_{se};Y;X^{(s)}) \ge I(Z^{(i)}_{se};Y;Z^{(s)}) $ holds due to data processing inequality ($X^{(s)}\rightarrow Z^{(s)}$). On the other side, $I(Z^{(i)}_{se};Y;X^{(s)})$ can be factorized into two terms by introducing $Z^{(s)}$:
\begin{equation}
    \begin{split}
        I(Z^{(i)}_{se};Y;X^{(s)})=I(Z^{(i)}_{se};Y;X^{(s)}|Z^{(s)})+I(Z^{(i)}_{se};Y;X^{(s)};Z^{(s)}).\label{a1}
    \end{split}
\end{equation}
For the first term of RHS in Eq.~\eqref{a1}:
\begin{equation}
    \begin{split}
        I(Z^{(i)}_{se};Y;X^{(s)}|Z^{(s)})=& I(Y;X^{(s)}|Z^{(s)})-I(Y;X^{(s)}|Z^{(i)}_{se},Z^{(s)}) \\
        =&0-I(Y;X^{(s)}|Z^{(i)}_{se},Z^{(s)})\le 0,
    \end{split}
\end{equation}
where $I(Y;X^{(s)}|Z^{(s)})=0$ using the definition of \textit{sufficiency}; For the second term of RHS in Eq.~\eqref{a1}:
\begin{equation}
    \begin{split}
        I(Z^{(i)}_{se};Y;X^{(s)};Z^{(s)})=I(Z^{(i)}_{se};Y;Z^{(s)})-I(Z^{(i)}_{se};Y;Z^{(s)}|X^{(s)}).\label{a3}
    \end{split}
\end{equation}
For the second term of RHS in Eq.~\eqref{a3}:
\begin{equation}
    \begin{split}
        I(Z^{(i)}_{se};Y;Z^{(s)}|X^{(s)})&=I(Y;Z^{(s)}|X^{(s)})-I(Y;Z^{(s)}|X^{(s)},Z^{(i)}_{se}) \\
        &=0-0=0,\label{a4}
    \end{split}
\end{equation}
where $I(Y;Z^{(s)}|X^{(s)})\!=\!I(Y;Z^{(s)}|X^{(s)},Z^{(i)}_{se})\!=\!0$ as $Z^{(s)}$ is a representation of $X^{(s)}$ using \textit{Hypothesis} 1. Therefore, combining  Eq.~\eqref{a1} - \eqref{a4}  concludes: \begin{equation}
    I(Z^{(i)}_{se};Y;X^{(s)})=I(Z^{(i)}_{se};Y;Z^{(s)}).
\end{equation}
\end{proof}
\end{theorem}

Following \textbf{Theorem 1}, and using \textit{Hypothesis} 3 and 4, we have:
\begin{equation}
I(Z^{(i)}_{se};Y;X^{(s)})=I(Z^{(i)}_{se};Y;Z_{sr}^{(i)})\le I(Z^{(i)}_{se};Z_{sr}^{(i)})\thickapprox 0.    
\end{equation}
Based on the above analysis, it is concluded that  $I(Z_{se}^{(i)};Y;X^{(s)})$ can be upper-bounded by $I(Z^{(i)}_{se};Z_{sr}^{(i)})\thickapprox 0$.
\section{On Loss Functions}
\label{B}
\setcounter{equation}{0}

In Section 3, we maximize mutual information between representation and label by minimizing cross-entropy loss (Eq. (5), (7) in the paper). We formulate this approximation as the following \textbf{Proposition 1}.
\begin{proposition}\label{mice} Let $X$ and $Y$ be random variables with domain $\mathcal{X}$ and $\mathcal{Y}$, respectively. Let $Z$ be a representation of $X$. Then, maximizing $I(Z;Y)$ can be approximated by minimizing cross-entropy loss of $q(y|z)$ given observations from $P(X,Y)$ as  $\{x_j,y_j\}_{j=1}^{N}$. $q(y|z)$ is regarded as classifier in practical.

\begin{proof}
Using the definitions of mutual information and entropy:
\begin{equation}
\max I(Z;Y) = H(Y) - H(Y|Z),
    \label{eq:mi}
\end{equation}
and as $H(Y)$ will not change if domain $\mathcal{Y}$ does not change, maximizing $I(Z;Y)$ is equivalent to minimizing $H(Y|Z)$:
\begin{equation}
\begin{split}
 \min H(Y|Z)=&\int  p(z)H(Y|Z=z) dz \\
       =&-\int\!\!\!\int p(z) p(y|z) \log p(y|z) \ dydz.
\end{split}
\end{equation}
As $D_{KL}(p(y|z)\|q(y|z))\!=\!\int p(y|z) \log p(y|z) - p(y|z) \log q(y|z) dz\!\ge\! 0$ holds identically:
\begin{equation}
\begin{split}
 \min  H(Y|Z)=&-\int\!\!\!\int p(z) p(y|z) \log p(y|z) \ dydz \\
 \le& -\int\!\!\!\int p(z) p(y|z) \log q(y|z) \ dydz \\
  =& -\int\!\!\!\int p(z,y) \log q(y|z) \ dydz \\
  =& -\int\!\!\!\int\!\!\!\int p(y|x)p(z|x)p(x) \log q(y|z) \ dxdydz. \\
\end{split}
\end{equation}
The last equation holds for $Z$ is conditional independent from $Y$ given X based on the graphical model illustrated in Section 3 in the paper ($Y\rightarrow X \rightarrow Z$), \ie,  $p(y,z|x)\!=\!p(y|x)p(z|x)$. For specific observations $\{x_j,y_j\}_{j=1}^{N}$ (and note that $p(z|x)$ is usually represented as a deterministic function), we can approximate the upper bound of $H(Y|Z)$ by Monte Carlo sampling:
\begin{equation}
\begin{split}
 \min  H(Y|Z)\le& -\int\!\!\!\int\!\!\!\int p(y|x)p(z|x)p(x) \log q(y|z) \ dxdydz \\
 \thickapprox& -\frac{1}{N} \sum_{j=1}^{N} \log q(y_j|z_j),
\end{split}
\end{equation}
which is a typical form of cross-entropy loss. Therefore, \textbf{Proposition 1} holds.
\end{proof}
\end{proposition}
\begin{remark}

For \textit{Hypothesis} 1 in Appendix~\ref{A}, if the approximation in \textbf{Proposition 1} is close enough, then we can infer that $D_{KL}(p(y|x)\|q(y|z))\!\rightarrow\!0^+$, which indicates the \textit{sufficiency} of $Z^{(s)}$ of $X^{(s)}$ for $Y$.
\end{remark}

In Section 3, we minimize cross-view conditional mutual information  by minimizing cross-entropy loss (Eq. (10), (11), (14), (15), (17) in the paper). We formulate this approximation as the following \textbf{Proposition 2}.
\begin{proposition}
Let $X^{(1)}$ and $X^{(2)}$ be random variables from visible modality and infrared modality (or generally two different views, \ie, modality view and body shape view), $Y$ be random variable of identity. Let $Z^{(1)}$ and $Z^{(2)}$  be representations of $X^{(1)}$ and $X^{(2)}$. Then minimizing $I(X^{(1)} ; Z^{(1)}| X^{(2)})$ can be approximated by minimizing cross-entropy between $p(y|z^{(2)})$ and $p(y|z^{(1)})$.
\begin{proof}
\begin{equation}
    \begin{split}
        I(X^{(1)} ; Z^{(1)}| X^{(2)})=&\int\!\!\!\int\!\!\!\int p(x^{(1)}, x^{(2)}, z^{(1)})\log \frac{p(z^{(1)},x^{(1)}|x^{(2)})}{p(z^{(1)}|x^{(2)})p(x^{(1)}|x^{(2)})}  \ dx^{(1)}dx^{(2)}dz^{(1)}  \\
        =&\int\!\!\!\int\!\!\!\int p(x^{(1)}, x^{(2)}, z^{(1)})\log \frac{p(z^{(1)}|x^{(1)},x^{(2)})p(x^{(1)}|x^{(2)})  }{p(z^{(1)}|x^{(2)})p(x^{(1)}|x^{(2)})}  \ dx^{(1)}dx^{(2)}dz^{(1)} \\
        =&\int\!\!\!\int\!\!\!\int p(x^{(1)}, x^{(2)}, z^{(1)})\log \frac{p(z^{(1)}|x^{(1)})  }{p(z^{(1)}|x^{(2)})}  \ dx^{(1)}dx^{(2)}dz^{(1)} \\
        =&\int\!\!\!\int\!\!\!\int p(x^{(1)}, x^{(2)}, z^{(1)})\log \frac{p(z^{(1)}|x^{(1)})p(z^{(2)}|x^{(2)})  }{p(z^{(1)}|x^{(2)})p(z^{(2)}|x^{(2)})}  \ dx^{(1)}dx^{(2)}dz^{(1)} \\
        =&\int\!\!\!\int p(x^{(1)}, x^{(2)}) D_{KL}(p(z^{(1)}|x^{(1)})\|p(z^{(2)}|x^{(2)})) \ dx^{(1)}dx^{(2)} \\
        &- \int p(x^{(2)}) D_{KL}(p(z^{(1)}|x^{(2)})\|p( z^{(2)}|x^{(2)})) \ dx^{(2)} \\
        \le&\int\!\!\!\int p(x^{(1)}, x^{(2)}) D_{KL}(p(z^{(1)}|x^{(1)})\|p(z^{(2)}|x^{(2)})) \ dx^{(1)}dx^{(2)}.
    \end{split}
\end{equation}
Thus, $ I(X^{(1)} ; Z^{(1)}| X^{(2)})$ can be upper-bounded by $D_{KL}(p(z^{(1)}|x^{(1)})\|p(z^{(2)}|x^{(2)}))$ integrated over $x^{(1)},x^{(2)}$.
We can approximate this KL divergence by:
\begin{equation}
\begin{split}
    D_{KL}(p(y|z^{(1)})\|p(y|z^{(2)}))=&\int p(y|z^{(1)})\log \frac{p(y|z^{(1)})  }{p(y|z^{(2)})} \ dy \\
=&\int p(y|z^{(1)})\log p(y|z^{(1)}) \ dy -\int p(y|z^{(1)})\log p(y|z^{(2)}) \ dy, \\
\end{split}
\end{equation}
where the first term of RHS of the last equation assumes to be constant, and the second term can be approximated by cross-entropy loss using Monte Carlo sampling similarly in \textbf{Proposition 1}. Therefore, \textbf{Proposition 2} holds.
\end{proof}

\end{proposition}

\section{Comparison to MPANet Using the Same Baseline}
\label{C}

We conduct an additional experiment to compare the performances of our method and others using our baseline. We choose MPANet~\cite{nuances}, which performed the highest accuracy on SYSU-MM01 among current open-source works. We reproduce it on our baseline. It is demonstrated in Table~\ref{tab:baseline} that our method achieves higher performances compared to MPANet using the same baseline.

\begin{table}[htbp]
  \centering
    \begin{tabular}{c|cc|cc|cc}
    \hline
    \multirow{3}*{Method} & \multicolumn{2}{c|}{\multirow{2}*{SYSU-MM01}} & \multicolumn{4}{c}{HITSZ-VCM} \\
\cline{4-7}          & \multicolumn{2}{c|}{} & \multicolumn{2}{c|}{Infrared-Visible} & \multicolumn{2}{c}{Visible-Infrared} \\
\cline{2-7}          & Rank-1 & mAP   & Rank-1 &  mAP  & Rank-1 &  mAP  \\
    \hline
    MPANet & 70.58 & 68.24 & 46.51 & 35.26 & 50.32 & 37.80 \\
    on our base & 71.39 & 67.77 &   58.46   &   45.69    &   61.01    & 46.98 \\
    \hline
    Ours  & 75.18 & 70.12 & 67.65 & 52.30  & 70.23 & 52.54 \\
    \hline
    \end{tabular}%
      \caption{Reproduce MPANet on our baseline. All Hyper-parameters have been carefully tuned.}
  \label{tab:baseline}%
\end{table}%

\end{onecolumn}

\end{document}